\documentclass[a4paper, 10pt]{article}
\usepackage[margin=2.5cm]{geometry}
\usepackage[utf8]{inputenc}
\usepackage[T1]{fontenc}

\usepackage{CJKutf8}
\newcommand{\zh}[1]{\begin{CJK*}{UTF8}{gbsn}#1\end{CJK*}}
\newcommand{\jp}[1]{\begin{CJK*}{UTF8}{min}#1\end{CJK*}}
\newcommand{\kr}[1]{\begin{CJK*}{UTF8}{mj}#1\end{CJK*}}
\usepackage{graphicx}
\graphicspath{{_figures/}}
\usepackage{hyperref}	
\hypersetup{colorlinks=true, linkcolor=blue, urlcolor=black, citecolor=blue}
\usepackage{booktabs}
\usepackage[labelfont=bf,labelsep=period,justification=justified, singlelinecheck=false]{caption}

\title{FineFreq: A  Multilingual Character Frequency Dataset from Web-Scale Text}
\author{Binbin XU}
\date{\normalsize{EuroMov Digital Health in Motion\\ Univ Montpellier, IMT Mines Ales, Ales, France}\\
{\texttt{binbin.xu@mines-ales.fr}}}

\begin{document}

\maketitle

\begin{abstract}
We present FineFreq, a large-scale multilingual character frequency dataset derived from the FineWeb and FineWeb2 corpora, covering over 1900 languages and spanning 2013-2025. The dataset contains frequency counts for 96 trillion characters processed from 57 TB of compressed text. For each language, FineFreq provides per-character statistics with aggregate and year-level frequencies, allowing fine-grained temporal analysis. The dataset preserves naturally occurring multilingual features such as cross-script borrowings, emoji, and acronyms without applying artificial filtering. Each character entry includes Unicode metadata (category, script, block), enabling domain-specific or other downstream filtering and analysis. The full dataset is released in both CSV and Parquet formats, with associated metadata, available on GitHub and HuggingFace. \url{https://github.com/Bin-2/FineFreq}
\end{abstract}

\section{Introduction}

Character frequency tables have long been fundamental in computational linguistics, text processing and human‑computer interaction  \cite{Ycart2012Letter,Cantone2003frequency}.
Early empirical work demonstrated that character occurrences in natural‑language texts are far from uniform, often following heavy‑tailed distributions rather than the simplistic equiprobable assumption \cite{Grigas2015Letter}. This non‑uniformity has deep consequences: many string‑matching algorithms, compression schemes, and encoding optimizations rely on accurate character‑frequency models rather than equi‑probable assumptions.

Beyond common algorithmic applications, character frequency can exert psychological and cognitive influence on reading and text comprehension. Studies on letter‑frequency effects show that more frequent letters are detected and processed faster in letter‑decision tasks \cite{New2011letter}. 
Frequency information is also useful and necessary in readability metrics, font design, and typographic optimization, especially in contexts such as on‑screen keyboards and constrained mobile‑device input where efficient character coverage matters \cite{Grigas2018Letter}. 

In multilingual settings, character frequency data becomes more valuable. Modern digital communication is inherently mixed‑script and cross‑language. For example, Latin acronyms may appear in non‑Latin languages, emoji and symbols extend beyond the conventional linguistic boundaries, and borrowed words or foreign‑script fragments become more and more common. 

At the same time, technologies such as multilingual tokenizers, optical character recognition (OCR) / handwritten text recognition (HTR) systems, font fallback engines, and input‑method editors require reliable, realistic character distributions to function robustly across languages and writing systems. This is especially true for CJK languages (Chinese, Japanese and Korean) where the OCR/HTR systems need to handle tens of thousands of characters. The output depends heavily on accurate and realistic character distributions to guide recognition, post‑processing, and error correction. \cite{Li2025Threshold, Nagata1998Japanese, Koiso2025Proposed, Le2025Training}.
Recent work in OCR has demonstrated that failing to account for distributional shifts across language and time can lead to errors, which has motivated researchers to incorporate the empirical frequency constraints into recognition pipelines \cite{Kaliosis2025Learning}.

Despite this broad and growing need, publicly available character-frequency resources remain limited. Most existing tables cover a small number of high‑resource languages or curated corpus, primarily in Latin or CJK scripts. 
One notable effort from 2012 computed character frequencies across 262 Wikipedia editions, covering approximately 18 billion characters and reporting the top 100 characters per language after various normalization steps \cite{Vrandecic2012How}. This 2012 dataset was very helpful at the time. However, modern multilingual NLP now operate at much larger scale and coverage.

Another important point is that most previous works offer only static, aggregated counts, even though languages evolve continuously. New symbols appear, writing habits shift, borrowed words come and go, and even basic character usage can move slowly from year to year. Because of this, having some temporal resolution is useful and sometimes necessary to understand how characters are really used in modern web text.

There is thus a lack of large‑scale, multilingual, temporally resolved character‑frequency data. This absence presents a bottleneck to research and development in multilingual NLP, cross‑script typography, dynamic font and input‑method optimization, and sociolinguistics.

To address this gap, we introduce {FineFreq}, a large-scale character frequency dataset covering over 1900 languages from 2013 to 2024/2025, derived from the full text of FineWeb \cite{Welbl2021Challenges,Penedo2023RefinedWeb,Penedo2024FineWeb} and FineWeb2 \cite{Penedo2025FineWeb2} which are two of the most extensive public web-scale multilingual corpora available. 
FineFreq provides per-language character frequency tables, including both aggregate counts and year-level distributions, allowing analyses of diachronic trends and modern usage patterns. 
The dataset contains more than 96 trillion characters, making it, to our knowledge, the most extensive multilingual character frequency resource to date. Each language's data is stored in both CSV and Parquet format, and accompanied by metadata and statistics for convenient access and inspection, accessible on GitHub and HuggingFace. 
This resource aims to support research and development in multilingual NLP, text rendering, corpus linguistics, and other areas that require detailed, up-to-date, and broad-coverage character distributions.

\section{Data and Method}

The sources of this work are FineWeb \cite{Welbl2021Challenges,Penedo2023RefinedWeb,Penedo2024FineWeb} and FineWeb2 \cite{Penedo2025FineWeb2}, two large-scale multilingual web text corpora. 
FineWeb provides English data extracted from Common Crawl dumps between 2013 and 2025. In total, 110 dumps totaling 48.58 TB (compressed) were used from version 1.4, from \texttt{CC-MAIN-2013-20} to \texttt{CC-MAIN-2025-26}. 
FineWeb2 (v2.1.0) provides preprocessed text across over 1900 languages, primarily covering the same period (2013-2024). This portion contributes an additional 8.58 TB, organized by language and filtered by quality and script consistency.
The combined source corpus exceeds 57 TB of compressed multilingual data.

\begin{figure}[!htb]
    \centering
    \includegraphics[width=\linewidth]{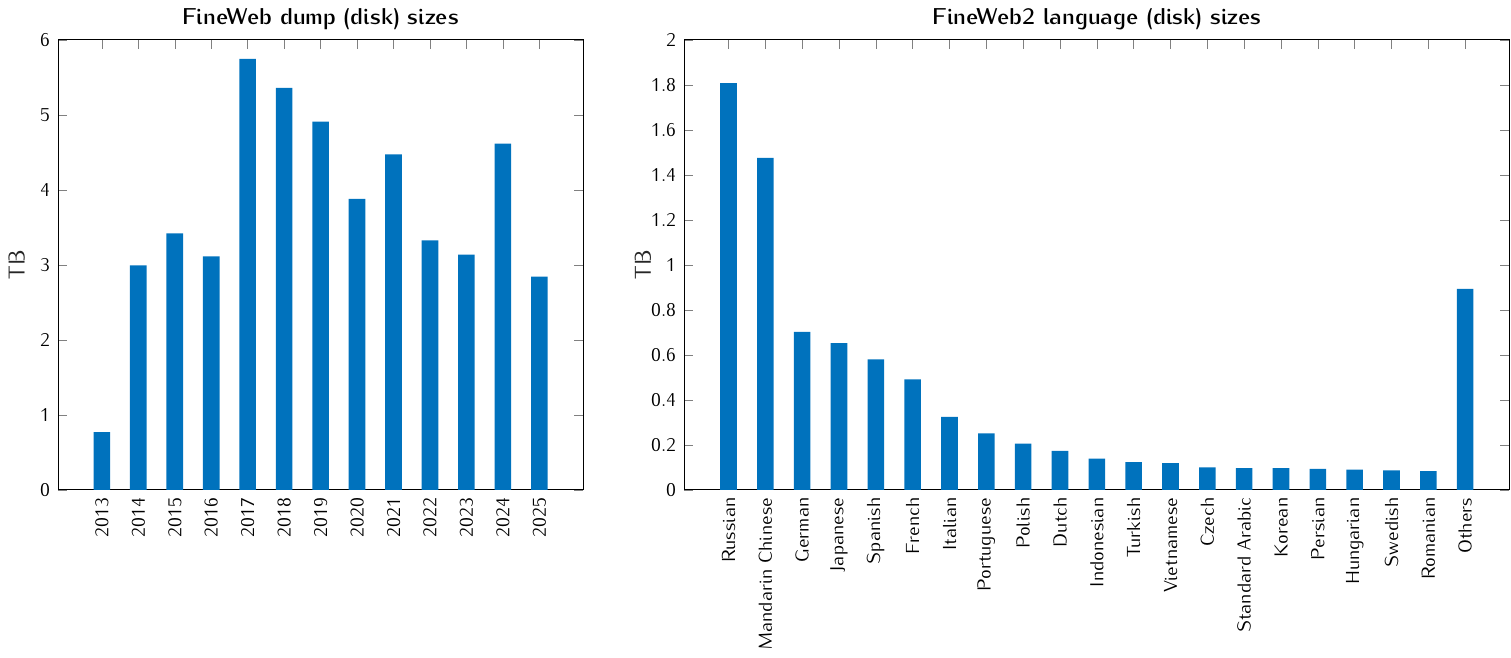}
    \caption{Left: Yearly disk sizes of FineWeb dumps (English only), v1.4.0. Right: Disk sizes of FineWeb2 multilingual data by top contributing languages, v2.1.0. In total 57.16 TB.}
    \label{fig:fineweb_sizes}
\end{figure}

Figure~\ref{fig:fineweb_sizes} illustrates the disk usage for these two corpora. On the left, we show the annual size of FineWeb dumps, with peak volumes in 2017-2019. On the right, the FineWeb2 data is sorted by size, where Russian, Mandarin Chinese, German, Japanese and Spanish dominate, each contributing more than 500 GB. The long-tail structure of the multilingual set is visible: although a few major languages account for large volumes (top 20 languages represent 7.68 TB), hundreds of smaller languages are represented at much lower volumes (in total, about 900GB).

During preprocessing, all text lines were normalized using Unicode Normalization Form Compatibility Composition (NFKC) to unify equivalent character sequences (e.g., full-width to ASCII forms, compatibility ligatures). All control characters in the \texttt{C0} (U+0000--U+001F) and \texttt{C1} (U+007F--U+009F) ranges were removed. These include standard non-printable characters such as \texttt{\textbackslash n}, \texttt{\textbackslash t}, and other legacy control codes. 

No lowercasing or case folding was applied during preprocessing. All character counts are case-sensitive by default, preserving the original orthographic distribution. Users may apply case normalization as needed for downstream applications.

To avoid biasing counts with repeated metadata, boilerplate, or duplicated content, we relied on the cleaned versions of FineWeb and FineWeb2, which already applied deduplication and quality filtering. No further filtering was used at the language or script level in this work. This decision reflects both practical and linguistic constraints: large-scale multilingual corpora naturally include borrowings, acronyms, emoji, and cross-script tokens due to code-switching and globalization. And Unicode codepoints do not cleanly map to language boundaries, especially for CJK and many scripts intentionally coexist (e.g., Latin acronyms in Japanese). Heuristic or lexicon-based filtering would likely introduce more bias than it eliminates. To support selective downstream filtering, we include Unicode metadata (script, category, block) for every character.

Given the scale of the input corpus (57 TB of compressed Parquet files, exceeding 200 TB uncompressed), all processing was performed in streaming mode, without loading full datasets into memory. The processing was executed on a  workstation equipped with dual Intel Xeon 6130 CPUs and 128 GB ECC RAM. A parallelized pipeline was implemented using 10 worker threads, each reading preprocessed Parquet shards line-by-line and computing character statistics in isolation. To overcome disk throughput bottlenecks, raw data was staged from HDD to SATA SSD via asynchronous background threads, while active counting processes operated exclusively on the SSD to maximize IO efficiency. This pipeline enabled complete corpus transfer and aggregation over several weeks of processing.

After per-line character counts were computed, statistics were grouped by character and aggregated across years using in-memory dictionaries. These dictionaries were then serialized into structured tables including both total and per-year frequency fields, preserving temporal resolution for each language.

\section{Dataset Description}

The released dataset contains character-level statistics for 1900+ languages, processed from over 96.58 trillion characters of multilingual web text. Each language's data includes both aggregate and yearly frequency tables, covering the period from 2013 and 2025, depending on the source. All resources are provided in CSV and Parquet formats, and accompanied by metadata and per-language statistics.

\begin{table}[h]
  \centering
  \caption{Top 20 languages by total character frequency.}
  \small
    \begin{tabular}{cccccrc}
    \toprule
    \textbf{ISO639-3} & \textbf{Lang Code} & \textbf{Script} & \textbf{Language} & \textbf{Chars} & \multicolumn{1}{c}{\textbf{Frequency}} & \textbf{Year Range} \\
    \midrule
    \texttt{eng}   & \texttt{eng\_Latn} & \texttt{Latn}  & English & 69\,211 & 81\,075\,397\,566\,715 & 2013-2025 \\
    \texttt{rus}   & \texttt{rus\_Cyrl} & \texttt{Cyrl}  & Russian & 52\,605 & 3\,520\,270\,440\,986 & 2013-2024 \\
    \texttt{deu}   & \texttt{deu\_Latn} & \texttt{Latn}  & German & 37\,599 & 1\,620\,982\,694\,281 & 2013-2024 \\
    \texttt{spa}   & \texttt{spa\_Latn} & \texttt{Latn}  & Spanish & 31\,018 & 1\,412\,270\,994\,363 & 2013-2024 \\
    \texttt{fra}   & \texttt{fra\_Latn} & \texttt{Latn}  & French & 35\,233 & 1\,170\,854\,379\,113 & 2013-2024 \\
    \texttt{cmn}   & \texttt{cmn\_Hani} & \texttt{Hani}  & Mandarin Chinese & 81\,717 & 946\,699\,852\,091 & 2013-2024 \\
    \texttt{ita}   & \texttt{ita\_Latn} & \texttt{Latn}  & Italian & 27\,894 & 779\,844\,597\,230 & 2013-2024 \\
    \texttt{por}   & \texttt{por\_Latn} & \texttt{Latn}  & Portuguese & 26\,843 & 590\,002\,142\,721 & 2013-2024 \\
    \texttt{jpn}   & \texttt{jpn\_Jpan} & \texttt{Jpan}  & Japanese & 50\,425 & 584\,060\,807\,797 & 2013-2024 \\
    \texttt{pol}   & \texttt{pol\_Latn} & \texttt{Latn}  & Polish & 23\,994 & 437\,026\,889\,016 & 2013-2024 \\
    \texttt{nld}   & \texttt{nld\_Latn} & \texttt{Latn}  & Dutch & 21\,548 & 422\,952\,993\,243 & 2013-2024 \\
    \texttt{ind}   & \texttt{ind\_Latn} & \texttt{Latn}  & Indonesian & 23\,064 & 371\,467\,212\,058 & 2013-2024 \\
    \texttt{tur}   & \texttt{tur\_Latn} & \texttt{Latn}  & Turkish & 21\,091 & 277\,878\,739\,727 & 2013-2024 \\
    \texttt{vie}   & \texttt{vie\_Latn} & \texttt{Latn}  & Vietnamese & 26\,943 & 259\,687\,322\,320 & 2013-2024 \\
    \texttt{swe}   & \texttt{swe\_Latn} & \texttt{Latn}  & Swedish & 17\,030 & 208\,965\,742\,427 & 2013-2024 \\
    \texttt{ces}   & \texttt{ces\_Latn} & \texttt{Latn}  & Czech & 19\,443 & 198\,755\,507\,561 & 2013-2024 \\
    \texttt{hun}   & \texttt{hun\_Latn} & \texttt{Latn}  & Hungarian & 19\,706 & 195\,272\,851\,661 & 2013-2024 \\
    \texttt{ron}   & \texttt{ron\_Latn} & \texttt{Latn}  & Romanian & 18\,124 & 191\,117\,019\,386 & 2013-2024 \\
    \texttt{fas}   & \texttt{fas\_Arab} & \texttt{Arab}  & Persian & 16\,509 & 183\,581\,646\,234 & 2013-2024 \\
    \texttt{nob}   & \texttt{nob\_Latn} & \texttt{Latn}  & Norwegian Bokmå & 17\,185 & 180\,325\,792\,915 & 2013-2024 \\
    \midrule
    \multicolumn{4}{r}{\textbf{Total (All Languages)}} &   & \textbf{96.58T} \\
    \bottomrule
    \end{tabular}
  \label{tab:top20_languages}
\end{table}

Table~\ref{tab:top20_languages} lists the top 20 languages ranked by total character count, showing their corresponding ISO codes, scripts, and cumulative frequency. English, drawn from FineWeb v1.4.0, dominates the dataset with over 81 trillion characters. FineWeb2 v2.1.0 contributes the rest, covering high-resource languages like Russian, German, Spanish, Mandarin Chinese, and Japanese. The coverage reflects realistic distributions of online multilingual content and writing systems.

\begin{table}[!htb]
\centering
\caption{Fields included in each per-language CSV and Parquet file.}
\small
\begin{tabular}{lll}
\toprule
\textbf{Field} & \textbf{Type} & \textbf{Description} \\
\midrule
\texttt{iso\_639\_3} & string & ISO 639-3 language code (e.g., \texttt{eng}) \\
\texttt{script} & string & ISO 15924 script code (e.g., \texttt{Latn}, \texttt{Cyrl}, \texttt{Hani}) \\
\texttt{language\_code} & string & Combined code of language and script (e.g., \texttt{eng\_Latn}) \\
\texttt{language\_name} & string & Human-readable name of the language (e.g., English) \\
\texttt{character} & string & A single Unicode character \\
\texttt{unicode\_category} & string & Unicode General Category (e.g., \texttt{Lu}, \texttt{Nd}, \texttt{Po}) \\
\texttt{unicode\_name} & string & Full Unicode name (e.g., \texttt{LATIN CAPITAL LETTER A}) \\
\texttt{total\_frequency\_all\_time} & int64 & Aggregated character count across all years \\
\texttt{time\_periods\_with\_data\_count} & int & Number of years where this character appears \\
\texttt{time\_periods\_list} & list & List of years where the character has nonzero frequency \\
\texttt{year\_20xx\_frequency} & int64 & Yearly frequency count for each year from 2013 to 2025 \\
\bottomrule
\end{tabular}
\label{tab:dataset_structure}
\end{table}

Each per-language file in the dataset (in both CSV and Parquet formats) contains frequency statistics for all characters observed in that language, along with metadata such as Unicode properties and yearly counts. Table~\ref{tab:dataset_structure} summarizes the full structure. In addition to total frequency, the dataset includes temporal information per-year counts from 2013 to 2025. This allows fine-grained diachronic analysis. Characters that are absent in a given year are assigned zero. The dataset is structured for ease of programmatic usability, with consistent field names and schemas across all files.

In addition to per-language character frequency tables, the dataset includes a manifest directory containing two index files: \texttt{languages.csv} and \texttt{languages.parquet}. These files summarize key metadata for each language included in the corpus. Each row corresponds to a unique language-script pair and provides information such as ISO 639-3 code, script, total character count, total frequency, years covered, and relative file path to the corresponding data folder.

\subsection*{Cross‑Language Character Frequency Profiles}

\begin{figure}[h]
    \centering
    \includegraphics[width=\textwidth]{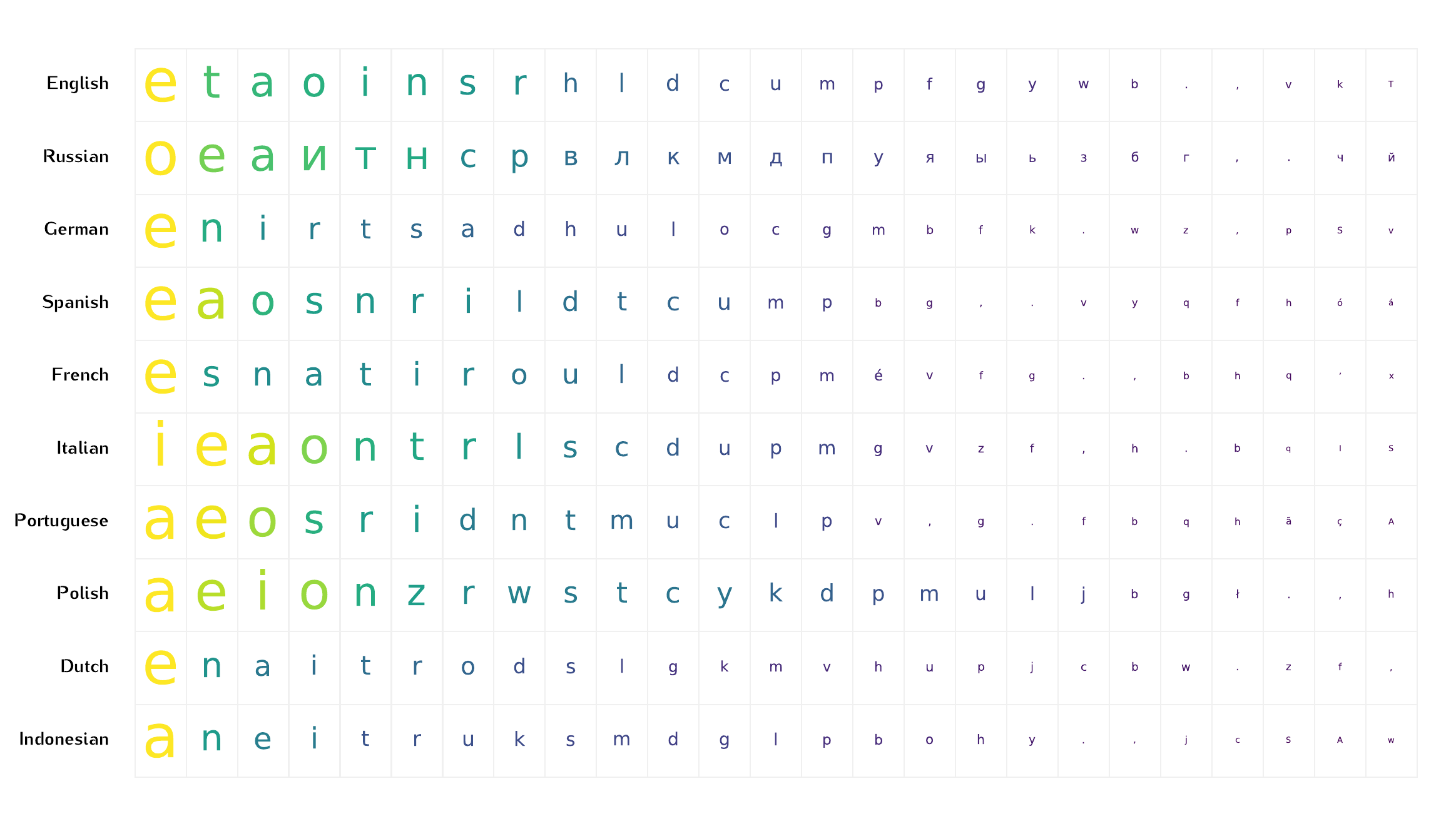}
    \caption{Top 25 characters by relative frequency for the top 10 non-CJK languages in the corpus. }
    \label{fig:char_rank}
\end{figure}

Figure~\ref{fig:char_rank} presents the top 25 characters by relative frequency for the 10 most frequent non-CJK languages in the dataset. Each row corresponds to a language, with characters sorted left-to-right by descending frequency. Font size and color are scaled by each character's normalized frequency within that language.
Accented characters, punctuation marks, and script-specific symbols (e.g., \texttt{'é'}, \texttt{'ñ'}, \texttt{'ł'}, \texttt{'ç'}, \texttt{'ü'}, etc.) naturally appear in the rankings, reflecting their real usage in the source corpora without filtering or normalization. This confirms that the frequency tables capture the actual orthographic diversity present in modern multilingual web text.

\begin{figure}[htb]
    \centering
    \includegraphics[width=\textwidth]{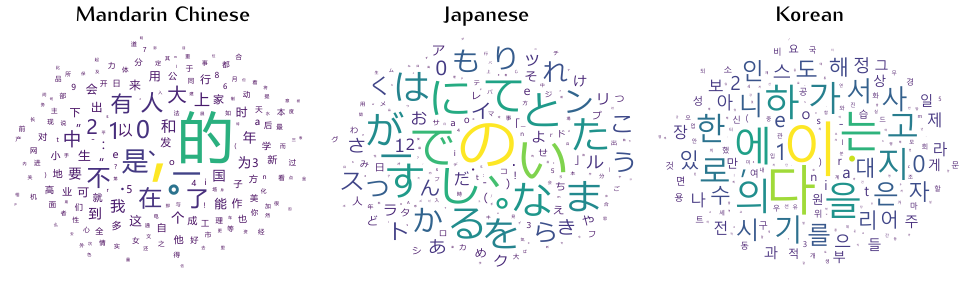}
    \caption{Character cloud (top 200) for CJK languages.}
    \label{fig:wc_cjk}
\end{figure}

Character clouds for Mandarin Chinese, Japanese, and Korean are shown in Figure~\ref{fig:wc_cjk}. The size and color of each character correspond to its relative frequency, normalized per language. These distributions capture the structural and linguistic patterns of each language.
In Chinese, the possessive and modifier marker ``\zh{的}" dominates, alongside frequent punctuation such as ``\zh{。}" and ``\zh{,}". Japanese shows the topic and genitive particle ``\jp{の}" as most frequent, followed by functional kana like ``\jp{い}", ``\jp{で}" and the punctuation mark ``\jp{、}". Korean exhibits high counts for grammatical markers such as ``\kr{이}", ``\kr{다}", ``\kr{는}" and ``\kr{에}", with the period ``." also appearing prominently. These clouds provide a structural snapshot of each language's orthographic core.
ASCII characters also appear in all three clouds, illustrating the integration of Latin-script tokens, digits, and punctuation into modern CJK web text.

\subsection*{Temporal Trends in Character Usage}

\begin{figure}[h]
    \centering
    \includegraphics[width=\textwidth]{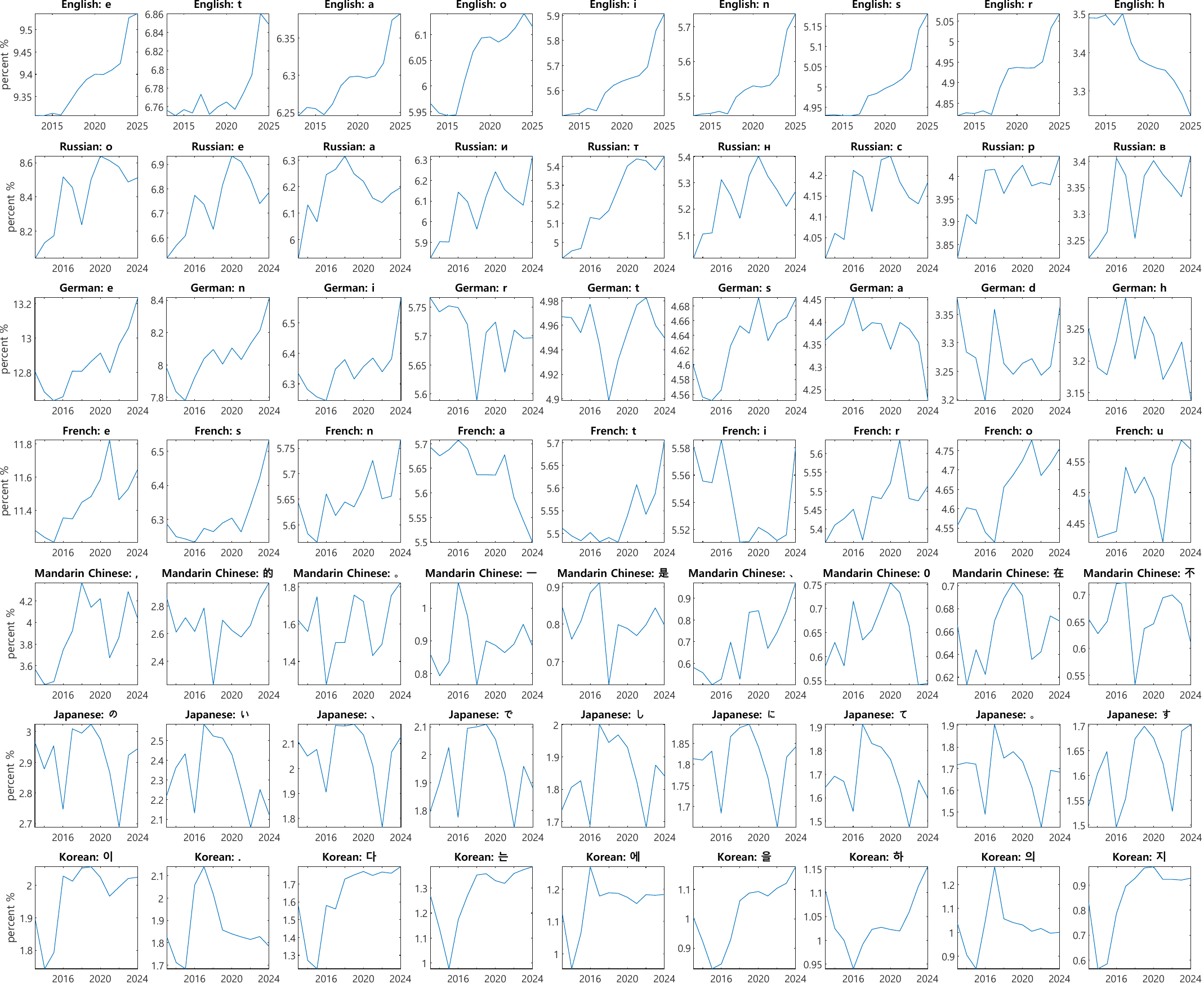}
    \caption{Yearly relative frequency of top 9 characters in 7 major languages. Each subplot shows the temporal trend of one character's proportion relative to all characters in that language for a given year.}
    \label{fig:top9_char_year}
\end{figure}

To illustrate the richness and resolution of the dataset, we plotted the normalized frequency of the top 9 non-space characters for 7 major languages (English, Russian, German, French, Mandarin Chinese, Japanese, and Korean), as shown in Figure~\ref{fig:top9_char_year}. 
These characters exhibit observable variation over time. For instance, the top 8 English characters generally increased in share after 2013, while \texttt{h} showed a marked decline. In Mandarin Chinese, characters fluctuate significantly year to year. Similar fluctuations appear in Hiragana and Hangul usage patterns in Japanese and Korean, respectively.

Although the percentage changes may appear visually small ($\leq 0.5\%$), they correspond to billions of characters. Such variation reflects natural shifts in linguistic usage, content sources, and web trends—changes that are largely invisible in smaller or static corpora.

\subsection*{Data availability}

All processed character‑frequency tables, metadata files are openly available in both CSV and Parquet formats through the project's GitHub repository: \url{https://github.com/Bin-2/FineFreq} .\\
A mirrored version is hosted on HuggingFace for programmatic access. 

\section{Conclusion}

This work introduces a large-scale, multilingual, and temporally resolved character frequency dataset constructed from 57 TB of cleaned web text spanning over 1900 languages. Built atop the FineWeb and FineWeb2 corpora, the dataset captures more than 96 trillion characters across thirteen years (2013-2025) and provides both aggregate and per-year frequency counts. The final release includes metadata, summary statistics, and character-level Unicode annotations to facilitate downstream analysis and reproducibility. All data is available in both CSV and Parquet formats through GitHub and Hugging Face.

Compared to earlier frequency tables derived from curated corpora such as newspaper, books or Wikipedia, this dataset reflects modern usage patterns drawn from the web at broader scale. The inclusion of year-level resolution offers a new perspective on orthographic shifts, lexical innovations, and script trends over time. Researchers can now study questions of typographic evolution, diachronic language change, and technical system design using frequency data based on trillions of real-world observations.

Considering that FineWeb and FineWeb2 were already carefully curated, this dataset makes no attempt to enforce more strict language boundaries or filter characters by predefined script lists. Given the scale and diversity of the web corpus, some level of cross-linguistic character use is inevitable. However, such sequences are not merely errors or noise. They represent the realities of modern multilingual communication, including code-switching, foreign word borrowing, and transliteration. These hybrid patterns reflect how digital language evolves in practice. By preserving this genuine variation, the dataset could be used in finer analysis for sociolinguistics, cross-lingual NLP, and the study of language interaction in multilingual environments. 

\bibliographystyle{ieeetr}
\bibliography{bib}

\begin{thebibliography}{10}

\bibitem{Ycart2012Letter}
B.~Ycart, ``Letter counting: a stem cell for cryptology, quantitative linguistics, and statistics,'' {\em arXiv preprint arXiv:1211.6847}, vol.~40,, pp.~3303--329, Nov. 2012.

\bibitem{Cantone2003frequency}
D.~Cantone and S.~Faro, ``On the frequency of characters in natural language texts,'' {\em TWLT 21 Algebraic Methods in Language Processing}, p.~69, 2003.

\bibitem{Grigas2015Letter}
G.~Grigas and A.~Juškevičienė, ``Letter frequency analysis of lithuanian and other languages using the latin alphabet,'' {\em Coactivity: Philology, Educology}, vol.~23, p.~81–91, Dec. 2015.

\bibitem{New2011letter}
B.~New and J.~Grainger, ``On letter frequency effects,'' {\em Acta Psychologica}, vol.~138, no.~2, pp.~322--328, 2011.

\bibitem{Grigas2018Letter}
G.~Grigas and A.~Juškevičienė, ``Letter frequency analysis of languages using latin alphabet,'' {\em ILR}, vol.~1, no.~1, pp.~p18--, 2018.

\bibitem{Li2025Threshold}
W.~Li, R.~M. Ramos, P.~C. Brom, and D.~L. Li, ``Threshold study for hanzi image recognition: Defining character and component limits in chinese, japanese, and korean script processing,'' {\em International Journal of Asian Language Processing}, vol.~35, no.~01, p.~2450011, 2025.

\bibitem{Nagata1998Japanese}
M.~Nagata, ``Japanese ocr error correction using character shape similarity and statistical language model,'' in {\em 36th Annual Meeting of the Association for Computational Linguistics and 17th International Conference on Computational Linguistics, Volume 2}, pp.~922--928, 1998.

\bibitem{Koiso2025Proposed}
N.~Koiso, Y.~Takemoto, Y.~Ishikawa, and M.~Takata, ``Proposed method of acquiring train data for early-modern japanese printed character recognizers,'' {\em The Journal of Supercomputing}, vol.~81, no.~6, pp.~764--, 2025.

\bibitem{Le2025Training}
A.~Le and A.~Kitamoto, ``Training kindai ocr with parallel textline images and self-attention feature distance-based loss,'' {\em arXiv preprint arXiv:2508.08537}, Aug. 2025.

\bibitem{Kaliosis2025Learning}
P.~Kaliosis and J.~Pavlopoulos, ``Learning to align: Addressing character frequency distribution shifts in handwritten text recognition,'' {\em arXiv preprint arXiv:2506.09846}, June 2025.

\bibitem{Vrandecic2012How}
D.~Vrandečić, ``How often is which letter?,'' 2012.

\bibitem{Welbl2021Challenges}
J.~Welbl, A.~Glaese, J.~Uesato, S.~Dathathri, J.~Mellor, L.~A. Hendricks, K.~Anderson, P.~Kohli, B.~Coppin, and P.-S. Huang, ``Challenges in detoxifying language models,'' {\em arXiv preprint arXiv:2109.07445}, Sept. 2021.

\bibitem{Penedo2023RefinedWeb}
G.~Penedo, Q.~Malartic, D.~Hesslow, R.~Cojocaru, A.~Cappelli, H.~Alobeidli, B.~Pannier, E.~Almazrouei, and J.~Launay, ``The refinedweb dataset for falcon llm: Outperforming curated corpora with web data, and web data only,'' {\em arXiv preprint arXiv:2306.01116}, June 2023.

\bibitem{Penedo2024FineWeb}
G.~Penedo, H.~Kydl{\'\i}{\v{c}}ek, L.~B. allal, A.~Lozhkov, M.~Mitchell, C.~Raffel, L.~V. Werra, and T.~Wolf, ``The fineweb datasets: Decanting the web for the finest text data at scale,'' in {\em The Thirty-eight Conference on Neural Information Processing Systems Datasets and Benchmarks Track}, 2024.

\bibitem{Penedo2025FineWeb2}
G.~Penedo, H.~Kydlíček, V.~Sabolčec, B.~Messmer, N.~Foroutan, A.~H. Kargaran, C.~Raffel, M.~Jaggi, L.~V. Werra, and T.~Wolf, ``Fineweb2: One pipeline to scale them all -- adapting pre-training data processing to every language,'' {\em arXiv preprint arXiv:2506.20920}, June 2025.

\end{thebibliography}
    
\end{document}